\documentclass[final,5p,times,twocolumn]{elsarticle}
\usepackage{amssymb, amsmath, amsthm, bm}
\usepackage[noend]{algpseudocode}
\usepackage{algorithmicx,algorithm}
\usepackage{graphicx}
\usepackage{caption}
\usepackage[colorlinks,linkcolor=blue]{hyperref}
\usepackage{epstopdf}
\usepackage[colorlinks,linkcolor=blue]{hyperref}
\usepackage{bbding}
\usepackage{multirow}
\usepackage{booktabs}
\usepackage{tikz,xcolor}

\newtheorem{rmrk}{Remark}
\numberwithin{equation}{section}

\journal{}

\begin{document}
\begin{frontmatter}
\title{Adversarial Momentum-Contrastive  Pre-Training}
\author[label1] {Cong Xu \fnref{cor1}}
\author[label1] {Dan Li \fnref{cor2}}
\author[label1] {Min Yang \corref{cor3}}
\fntext[cor1] {Email: congxueric@gmail.com }
\fntext[cor2] {Email: danliai@hotmail.com}
\cortext[cor3] {Corresponding author: yang@ytu.edu.cn}

\address[label1]{School of Mathematics and Information Sciences, Yantai University, Yantai 264005, China}

\begin{abstract}
Recently proposed adversarial self-supervised learning methods usually require big batches and long training epochs to extract robust features,
which will bring heavy computational overhead on platforms with limited resources.
In order to help the network learn more powerful feature representations in smaller batches and fewer epochs,
this paper proposes a novel adversarial momentum contrastive learning method, which introduces two memory banks
corresponding to clean samples and adversarial samples, respectively.
These memory banks can be dynamically incorporated into the training process to track invariant features among historical mini-batches.
Compared with the previous adversarial pre-training model, our method achieves superior performance with smaller batch size and less training epochs.
In addition, the model outperforms some state-of-the-art supervised defensive methods on multiple benchmark datasets after being fine-tuned on downstream classification tasks.
\end{abstract}

	
\end{frontmatter}

\section{Introduction}

Although deep neural networks (DNNs) have achieved state-of-the-art performance on many challenging computer vision tasks,
it was soon found that they are extremely vulnerable to semantic invariant corruptions \cite{geirhos2019aug,miko2018aug,rusak2020aug}
or adversarial attacks  \cite{goodfellow2015, szegedy2013},
which means that very small perturbations on the original input could cause the network to make wrong prediction.

To remedy this deficiency, robust feature learning has attracted the attention of researchers \cite{dong2022double,lecuyer2019defense}.
Among them,
supervised adversarial training \cite{goodfellow2015, madry2018pgd,tramer2019pc,zhang2019trades},
which guides the learning of neural networks with additional adversarial samples,
has become the most popular method.
Recently,
some unsupervised robust learning methods  \cite{chen2020rcl,jiang2020acl} have also been developed.
The main idea is to treat adversarial samples as a special kind of data augmentations,
and then adopt some self-supervised learning framework to extract the robust features contained in the data.
However, unlike ordinary data augmentations, which are usually determined before training,
adversarial examples are dynamically generated through training iterations and are greatly influenced by the current network parameters.
So in order to capture the invariant features among them,
very large batch sizes and long training epochs must be used \cite{chen2020rcl,jiang2020acl},
which will bring heavy computational overhead on platforms with limited resources.

In this work, we aim to address the above problem by developing a novel \textbf{A}dversarial \textbf{MO}mentum-\textbf{C}ontrastive (AMOC) learning approach,
which can leverage the historical information of mini-batches to help extract robust feature representations in a more efficient manner.
More specifically, we build two memory banks to track clean and adversarial feature representations that are consistent across different mini-batches, respectively.
The clean memory bank provides negative keys from clean data for contrastive learning, thus guaranteeing the natural accuracy of the model,
whereas the adversarial memory bank provides negative keys from historical adversarial examples,
thereby avoiding the learning inconsistency problem caused by the dynamic changes of the adversarial samples.

It is worthy to point out that in the original contrastive framework \cite{he2019moco} there is only one clean memory bank for standard training.
However, in adversarial training, we need not only clean samples but also their adversarial counterparts,
and the distributions of clean and adversarial samples are quite different,
so it is necessary to build separate memory banks for clean and adversarial samples, respectively.
With the aid of the sperate memory banks,
the network can better extract the invariant features contained in the perturbed data.
Our experimental results demonstrate that using separate memory banks does perform better than a single one.
Moreover, extensive experiments show that the proposed method even outperforms some state-of-the-art supervised defensive methods on multiple benchmark datasets
after being fine-tuned on downstream classification tasks.

The main contributions of this work are summarized as follows:
\begin{itemize}
	\item
    We give insights into why current adversarial contrastive learning algorithms require large batch sizes and long training epochs to reach convergence.
     Based on this, we explain the necessity of introducing separate memory banks to learn a better query encoder.
    \item
    We develop a novel contrastive learning framework named AMOC, which integrates a standard clean memory bank with an additional adversarial memory bank.
    And we also show how to train the framework efficiently in an end to end manner.
    \item
    We analytically identify a good form of training loss for AMOC and empirically discuss the  deficiencies of other loss forms,
    with particular emphasis on comparison with the single memory bank mode.
    \item
    Through extensive empirical evaluation, we show that AMOC is an effective defense strategy against a wide range of recently proposed state-of-the-art attacks in the literature.
    Compared with the existing contrastive pre-training methods, the proposed approach exhibits superior robustness and better computational efficiency.
    After fine-tuning on classification tasks, AMOC can successfully outperform many supervised adversarial defenses on the widely used datasets.
\end{itemize}

The remainder of the paper is organized as follows.
Section \ref{section-related-works} introduces the related works regarding adversarial robustness and contrastive learning.
Section \ref{section-major} elaborates the research motivation and introduces the framework of AMOC.
Section \ref{section-exp} gives a comprehensive  experimental results to demonstrate the efficiency of the proposed method.
Finally, the conclusion is drawn in Section 5.

\section{Related Works}
\label{section-related-works}

\subsection{Attack models}
Since the discovery that neural networks are vulnerable to artificial perturbations,
a series of attack methods have emerged to test the robustness of networks
\cite{carlini2017cwl2,croce2020aa,moosavi2016deepfool,madry2018pgd,szegedy2013,tramer2019pc}.
These attack models craft small perturbations to clean samples to generate various adversarial examples.
Fast gradient sign method (FGSM)  \cite{goodfellow2015} is a simple yet effective attack method that utilizes the sign of the gradients of the loss function to generate adversarial samples.
Projected gradient descent (PGD) \cite{madry2018pgd}  is a more powerful iterative attack that starts from a random position in the neighborhood of a clean input and then applies FGSM for several iterations.
DeepFool \cite{moosavi2016deepfool} is a gradient-based attack algorithm that iteratively linearizes the classifier
to generate the smallest perturbation sufficient to change the classification label.
C\&W \cite{carlini2017cwl2} is one of the most powerful attack to detect adversarial samples in $\ell_2$ norm.
Sparse $\ell_1$ descent (SLIDE) \cite{tramer2019pc} is a more efficient model that  overcomes the inefficiency of PGD in searching for $\ell_1$ perturbations.
Recently, Croce et al. \cite{croce2020aa} combined four diverse attacks into a more aggressive one, AutoAttack,
as a new benchmark for empirical robustness evaluation

\subsection{Adversarial training}
To enhance the robustness of neural networks against adversarial attacks, numerous defense methods have been developed from different perspectives
\cite{dong2022double,huang2021arch,lecuyer2019defense,rade2021hat,xu2022}.
Among them, adversarial training  \cite{goodfellow2015, madry2018pgd, zhang2019trades},
which minimizes the worst-case loss in the perturbation region, is considered to be one of the most powerful defenses.
For example, given a distribution $\mathcal{D}$  over samples $  x $ and labels $  y $,
standard adversarial training \cite{madry2018pgd} optimizes the following objective:
\begin{align*}
    \min_{\theta} \mathbb{E}_{(x,y) \in \mathcal{D}} \max_{\|\delta\| \le \epsilon} \mathcal{L}(x+\delta,y;\theta),
\end{align*}
while tradeoff-inspired adversarial defense (TRADES) \cite{zhang2019trades} considers a trade-off between natural accuracy and adversarial robustness:
\begin{align*}
     \min_{\theta} \mathbb{E}_{(x,y) \in \mathcal{D}} \left(\mathcal{L}(x,y;\theta)+\frac{1}{\lambda}\max_{\|\delta\| \le \epsilon} \mathcal{L}(x,x+\delta;\theta)\right).
\end{align*}

Nonetheless, adversarial training-based methods suffer from heavy computational overhead and are not scalable.
The generality and reusability of pre-trained models can alleviate this problem to some extent,
since robust models required for various downstream tasks can be obtained after simple fine-tuning.

\subsection{Contrastive learning}

Contrastive learning \cite{chen2020simclr,he2019moco,oord2019cpc}
is a popular self-supervised learning framework,
which maximizes the similarity of a sample to its distinct views and
minimizes its similarity with other instances.
Different contrastive learning approaches usually adopt different strategies to generate various views
and negative keys.
One of the simple yet efficient framework is SimCLR \cite{chen2020simclr},  which uses augmented views in the current mini-batch as negative keys.
On the other hand,  MoCo  \cite{he2019moco}  introduces an additional memory bank
to maintain negative representations from neighboring mini-batches.
It was found that contrastive learning can resist to semantic invariant corruptions due to its strong data augmentations \cite{hendrycks2019cifarc}.

\begin{table}[hbt]
	\caption{
		A short recapitulation of the related works.
	}
	\label{table-differences}
   	\centering
	\scalebox{0.55}{
	\begin{tabular}{ccc}
		\toprule
		& Related works & Major difference \\
		\midrule
		\multirow{2}*{Attack models} &\cite{goodfellow2015,madry2018pgd,tramer2019pc}  & gradient-based attacks to yield feasible adversarial perturbations \\
		&\cite{carlini2017cwl2,moosavi2016deepfool}  & optimization algorithms to find the ``minimal'' perturbations \\
		\midrule
		\multirow{3}*{Adversarial defenses} & \cite{dong2022double,kannan2018alp,madry2018pgd,zhang2019trades,rade2021hat} & supervised learning using adversarial samples\\
		&\cite{chen2020rcl,jiang2020acl,kim2020rocl,meng2021rp} & unsupervised learning using adversarial samples \\
		&\cite{lecuyer2019defense,xu2022} & defenses that do not use adversarial examples \\
		\midrule
		\multirow{2}*{Contrastive learning}
		& \cite{he2019moco} & w/ memory banks \\
		& \cite{chen2020simclr,oord2019cpc} & w/o memory banks \\
		\bottomrule
	\end{tabular}
	}
\end{table}

\subsection{Adversarial self-supervised pre-training}
Several recent works \cite{chen2020rcl,jiang2020acl,meng2021rp} began to treat the adversarial perturbation as a special data augmentation
and adopt self-supervised learning to obtain the robust feature representations from data.
The learned feature representations can further be transferred to various downstream tasks.
However, unlike ordinary data augmentations, which are usually determined before training,
adversarial examples are dynamically generated through training iterations and are greatly influenced by the current network parameters.
So in order to capture the invariant features among them,
very large batch sizes and long training epochs must be used.
Note that each adversarial sample requires several rounds of forward propagation and backward gradient calculations.
At this point, large batches and long training cycles will bring additional computational overhead far beyond ordinary contrastive learning.

To address the above problem, inspired by the work of \cite{he2019moco},
we develop a momentum-based adversarial pre-training framework, which consists of a standard clean memory bank and an additional adversarial memory bank.
The newly proposed adversarial bank aims to alleviate the inconsistency of adversarial samples in each iteration,
thereby reducing the learning difficulty and speeding up the training convergence.

\section{Adversarial Momentum-Contrastive Learning}
\label{section-major}

\subsection{Preliminaries}
We first recall the MoCo framework \cite{he2019moco} for learning  on clean data.
Let $ \mathcal{C} $ denote a set of  data augmentation operations.
For any augmentations $c, c' $ in $  \mathcal{C} $,
$(c(x), c'(x)) $ forms a pair of \textit{positive} augmented samples of $x$.
The model includes a query encoder $f_q $ and a key encoder $f_k $.
Denote by $  q=f_q(c(x)) $ the query encoder representation,
$ k_+=f_k(c'(x)) $ the \textit{positive key},  and $ k_- $ the \textit{negative key},
which could be the representation of any other sample from the key encoder.
The MoCo builds a memory bank $ \mathcal{M} $ to keep the negative keys of recent mini-batches.
Its optimization objective is then to minimize the following InfoNCE  loss:
\begin{align}
\label{MoCo}
\begin{split}
     &\mathcal{L}_{\mathrm{NCE}}(f_q(c(x)),f_k(c'(x)), \mathcal{M})\\[5pt]
         =&	-\log \frac{\exp(q \cdot k_+ / T)}{ \exp(q \cdot k_+ / T) + \sum_{k_- \in \mathcal{M}} \exp( q \cdot k_- / T) },
\end{split}
\end{align}
where $ T $ is a temperature constant.
During training, the mini-batch from the key encoder is subsequently enqueued into the memory bank $ \mathcal{M} $
and at the same time the oldest ones are dequeued.

From an intuitive point of view,
\eqref{MoCo} forces the feature representation of each sample in the query encoder to be consistent with its augmented sample,
and different from all other samples held in $\mathcal{M}$,
thereby helping the query encoder learn the invariant feature representations among the data.

The query encoder $ f_q$  can be learned by standard back propagation,
while the key encoder $ f_k $ is updated by the following rule \cite{he2019moco}:
\begin{align}
\label{momentum}
      \theta_k \leftarrow  m \theta_k + (1 - m) \theta_q,
\end{align}
where $m \in (0, 1) $ is a momentum coefficient, and $\theta_q$ and $\theta_k$ are parameters of query encoder and key encoder, respectively.

\subsection{AMOC pre-training framework}
\label{AMfr}

Recent works \cite{chen2020rcl,jiang2020acl} considered adversarial perturbations as a  special kind of data augmentations,
and used contrastive learning technique to extract the robust feature representations in the data.
However, adversarial examples are dynamically generated through training iterations and are greatly influenced by the current network parameters.
In order to capture the invariant features among them,
very large batch sizes and long training epochs must be used.
Since the calculation of each adversarial example requires heavy computational resources,
computational burden brought by large batches and long-term training can far exceed that in ordinary contrastive learning.

\begin{figure}[htb]
	\centering
	\begin{minipage}{0.4\textwidth}
		\centering
		\scalebox{0.3}{\includegraphics{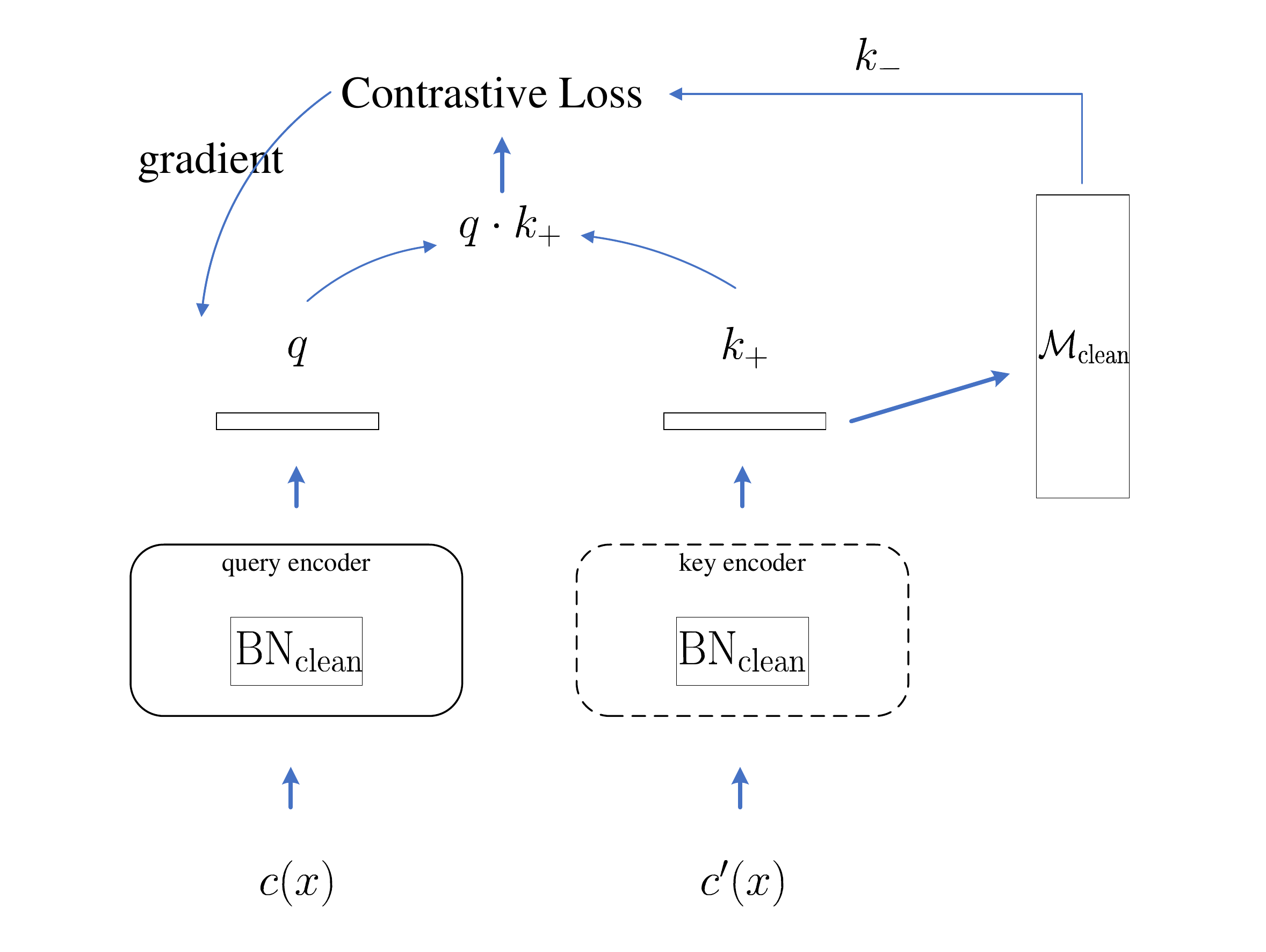}}
		\caption*{(a) MoCo}
	\end{minipage}
	\begin{minipage}{0.4\textwidth}
		\centering
		\scalebox{0.3}{\includegraphics{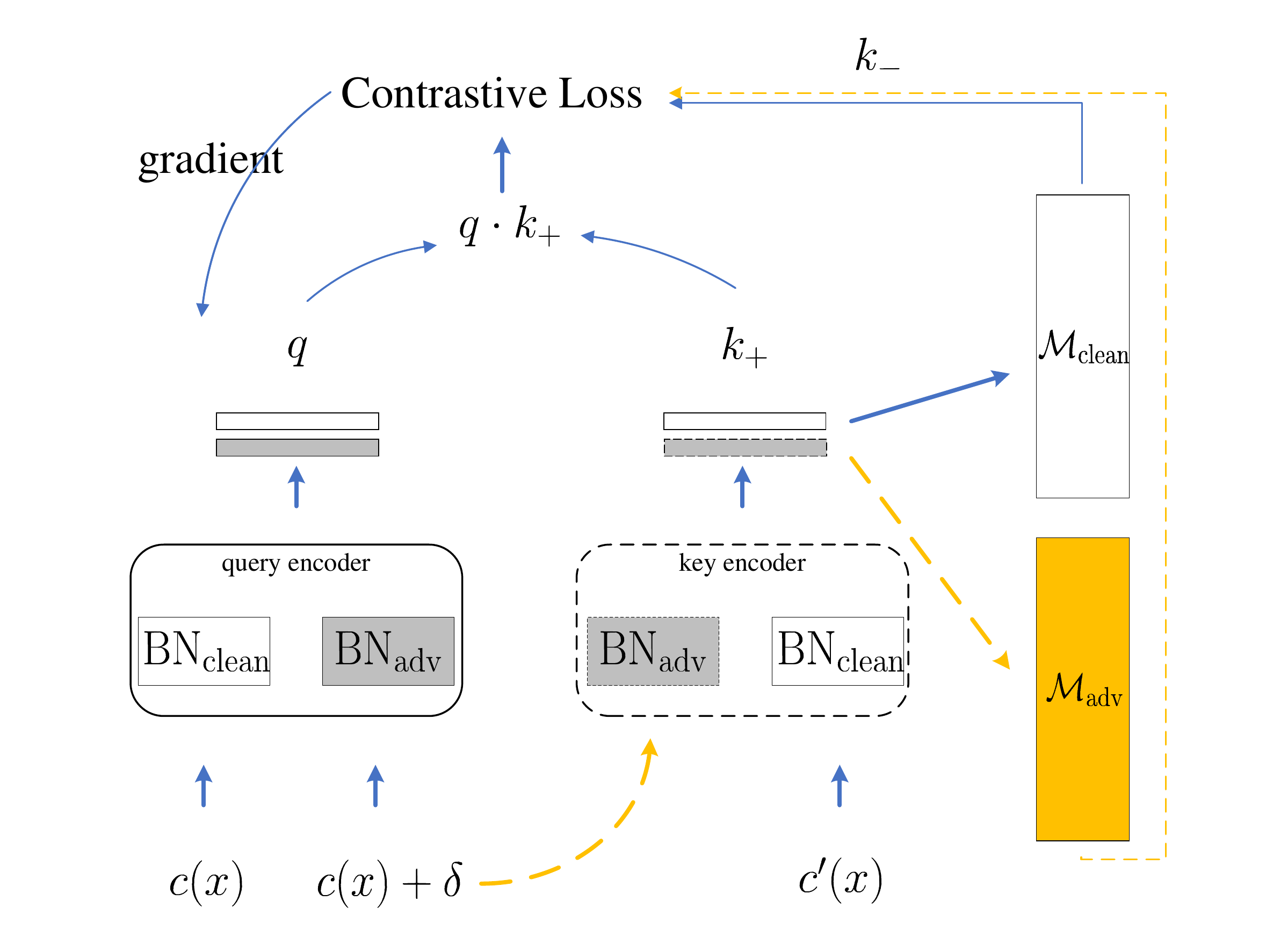}}
		\caption*{(b) AMOC}
	\end{minipage}
	\caption{
		(a) MoCo \cite{he2019moco} framework introduces the memory bank $\mathcal{M}_{\mathrm{clean}}$ for historical information.
		(b) AMOC framework utilizes two memory banks $\mathcal{M}_{\mathrm{clean}} $ and $ \mathcal{M}_{\mathrm{adv}} $ to maintain the historical clean and adversarial keys, respectively.
                  Here $c, c'$ denote distinct augmentations;
		         $\delta$ is the perturbation injected into the query;
		         $\mathrm{BN}_{\mathrm{clean}}, \mathrm{BN}_{\mathrm{adv}}$ are independent batch normalization modules for clean and adversarial samples;
		         $q, k_+$ is a pair of encoded feature representations from the same sample;
		         $k_-$ denotes the negative key from the memory banks.	}
	\label{framework}
\end{figure}

Inspired by the work of \cite{he2019moco},
we are to address the above problem by building two dynamic memory banks $\mathcal{M}_{\mathrm{clean}}$ and $\mathcal{M}_{\mathrm{adv}}$
to maintain clean and adversarial feature representations that are consistent across different mini-batches, respectively.
As in \cite{he2019moco}, the clean memory bank provides negative keys from clean data for contrastive learning, thus guaranteeing the natural accuracy of the model.
Whereas the newly proposed adversarial memory bank provides negative keys from historical adversarial samples,
thereby alleviating the inconsistency of adversarial samples and ensuring a faster convergence.

\begin{rmrk}
    Aside from the same clean memory bank $\mathcal{M}_{\mathrm{clean}}$ as in MoCo,
	AMOC introduces an additional adversarial memory bank $\mathcal{M}_{\mathrm{adv}}$ to maintain adversarial keys.
	However, note that there are only clean keys in the original key encoder
	In order to update the adversarial memory bank,
    after each iteration,
    we feed  the perturbed inputs of the query encoder to the key encoder (the dotted yellow lines in Figure \ref{framework}),
	and then $\mathcal{M}_{\text{adv}}$ is updated in a first-in and first-out fashion,
   that is, the new adversarial keys are enqueued into $\mathcal{M}_{\text{adv}}$ while the oldest ones are dequeued.
\end{rmrk}

In addition, since the distributions of adversarial and clean samples are  different,
we will utilize the dual Batch Normalization (BN) suggested in \cite{xie2020bn},
i.e., one BN for clean samples and the other for adversarial ones.
Then the corresponding encoder network is denoted as $ f(\cdot; \mathrm{BN_{clean}}) $ or $ f(\cdot; \mathrm{BN_{adv}}) $, respectively.

Now we give the formulation of the optimization objective of AMOC.
Most of all, a robust neural network should guarantee the feature representations of clean samples,
so the first part in our optimization objective is the standard MoCo loss \cite{he2019moco}:
\begin{align*}
\mathcal{L}_{\mathrm{CCC}}= \mathcal{L}_{\mathrm{NCE}}(f_q(c(x);\mathrm{BN_{clean}}), f_k(c'(x);\mathrm{BN_{clean}}), \mathcal{M}_{\mathrm{clean}}),
\end{align*}
where \textbf{CCC} means that \textbf{C}lean query encoder, \textbf{C}lean key encoder and \textbf{C}lean memory bank are used in this loss.

Furthermore, a robust neural network needs to be able to resist perturbations and learn invariant features under adversarial attacks.
To this end, we formulate the second loss as follows
\begin{align*}
\mathcal{L}_{\mathrm{ACA}}= \mathcal{L}_{\mathrm{NCE}}(f_q(c(x)+\delta;\mathrm{BN_{adv}}), f_k(c'(x);\mathrm{BN_{clean}}), \mathcal{M}_{\mathrm{adv}}).
\end{align*}
where \textbf{A}dversarial query encoder, \textbf{C}lean key encoder and \textbf{A}dversarial memory bank are used.
Note that in the above loss, we add adversarial perturbation to the input of the query encoder,
while leaving the input of the key encoder as it is.
In this way, the query encoder is forced to learn common feature representations contained in adversarial and clean samples,
i.e., robust features invariant under perturbations.

Combining the standard MoCo loss and the proposed adversarial loss,
we arrive at our  final optimization objective of the pre-training:

\begin{align}
	\mathcal{L} := \lambda \mathcal{L}_{\mathrm{CCC}} + (1- \lambda) \mathcal{L}_{\mathrm{ACA}},
 \label{FinalLoss}
\end{align}
where the hyperparameter $\lambda \in (0, 1)$  controls the effects of the clean samples and adversarial samples.

To minimize the objective \eqref{FinalLoss},
the parameter  $\theta_q$ of the query encoder could be learned via standard back propagation,
while the parameter $\theta_k$ should be updated by using \eqref{momentum},
otherwise  the key encoder will change rapidly and thus reduce the consistency of memory banks \cite{he2019moco}.
The entire optimization procedure of AMOC is summarized in Algorithm 1.

\begin{algorithm}[htb]
    \caption{Training procedure of AMOC}
    \begin{algorithmic}[1]
		\Require the image set $\{x\}$, augmentation operations $\mathcal{C}$,
		the query encoder $f_q(\cdot;\theta_q)$ and the key encoder $f_k(\cdot;\theta_k)$.
		\For{each mini-batch}
		\State Sample augmentations $c$ and $c'$ from $\mathcal{C}$ and craft adversarial perturbations $\delta$;
		\State Compute three encoded feature representations, respectively:
		\begin{align*}
		f_q\big(c(x);\mathrm{BN_{clean}}\big), \;  f_q \big(c(x)+\delta;\mathrm{BN_{adv}}\big), \; f_k \big (c'(x);\mathrm{BN_{clean}}\big);
		\end{align*}
		\State Calculate the loss \eqref{FinalLoss} and update the query encoder $f_q$ via back propagation;
        \State Update the key encoder $f_k$ via \eqref{momentum};
		\State Update the clean memory bank using $ f_k \big (c'(x);\mathrm{BN_{clean}}\big)$ and adversarial memory bank using  $f_k \big(c(x)+\delta;\mathrm{BN_{adv}}\big)$;
		\EndFor
		\Ensure the query encoder $f_q(\cdot;\mathrm{BN_{adv}})$.
    \end{algorithmic}
\end{algorithm}

\section{Experiment}
\label{section-exp}

\subsection{Experimental setup}

\textbf{Datasets.} In this section, we evaluate the performance of our approach on three commonly used datasets,
CIFAR10 \cite{krizhevsky2009cifar}, CIFAR100 \cite{krizhevsky2009cifar} and  CIFAR10-C \cite{hendrycks2019cifarc}.
Both of CIFAR10 and CIFAR100 contain 60000 $32 \times 32$ color images, which are divided into a training set of 50000 images
and a test set of 10000 images, while CIFAR10-C consists of nineteen different types of semantic invariant corruptions.
We use CIFAR10 and CIFAR100 to evaluate the adversarial robustness and CIFAR10-C  to measure the persistence against common corruptions.

\textbf{Attacks.} To reliably evaluate the adversarial robustness of defense methods,
several adversarial attacks including
PGD \cite{madry2018pgd}, DeepFool \cite{moosavi2016deepfool}, C\&W \cite{carlini2017cwl2},
SLIDE \cite{tramer2019pc} and AutoAttack \cite{croce2020aa} are adopted.
All implementations of them are provided by FoolBox \cite{foolbox2017} except AutoAttack from the source code of \cite{croce2020aa}.
The default settings of these attacks are listed in Table \ref{table-settings},
wherein step size denotes the relative step size of PGD and SLIDE,
the learning rate of C\&W, and the overshoot of DeepFool, respectively.
For brevity,  denote by PGD20 the shorthand of PGD with 20 iterations.

\begin{table}[htb]
	\caption{The basic setup for adversarial attacks in $\ell_{\infty}$, $\ell_{1}$ and $\ell_{2}$ norms.}
	\label{table-settings}
	\centering
	\scalebox{0.55}{
		\begin{tabular}{ccccccccccc}
			\toprule
			& \multicolumn{4}{c}{$\ell_{\infty}$, $\epsilon=8/255$} & & \multicolumn{2}{c}{$\ell_1$, $\epsilon=12$} & & \multicolumn{2}{c}{$\ell_{2}$, $\epsilon=0.5$} \\
			\cmidrule{2-5} \cmidrule{7-8} \cmidrule{10-11}
			& PGD & PGD & DeepFool& AutoAttack & & PGD & SLIDE &  & PGD & C\&W  \\
			\midrule
			number of iterations & 10 & 20 & 50  & - & & 50 & 50 & & 50&  1000 \\
			step size & 0.25 & 0.1 & 0.02& - & &0.05 & 0.05  &  & 0.1 & 0.01  \\		
			\bottomrule
	\end{tabular}}
\end{table}

\textbf{Baselines.}
The quality of learned feature representations is of most interest to pre-training methods,
which can be evaluated by a linear classifier using the pre-trained feature representations.
We choose adversarial contrastive learning (ACL)  \cite{jiang2020acl} as the baseline comparison,
which is a state-of-the-art self-supervised framework without memory banks.

We also compare  AMOC with several benchmark supervised defenses including
PGD adversarial training (PGD-AT) \cite{madry2018pgd}, adversarial logits pairing (ALP) \cite{kannan2018alp} and TRADES \cite{zhang2019trades}.
These defenses are implemented following the default settings of original papers.
The adversarial samples required in the procedure of training
are all crafted by PGD10 attack within $\epsilon=8/255$.

\textbf{Hyperparameters.}
There are three hyperparameters in our framework, including the momentum of $m$, the temperature of $T$ and the length of $K$.
We set $m=0.999$ and $T=0.2$ as suggested in \cite{he2019moco} but use a smaller $K=32768$ instead.
The weight parameter in \eqref{FinalLoss} is chosen as $\lambda=0.5$   to balance the two loss terms (see Section \ref{section-options} for sensitivity analysis).

\textbf{Implementation details.}
We take the ResNet-18 \cite{he2016resnet} as the backbone
and apply 5-step and 10-step $ l_\infty $  PGD attack  \cite{madry2018pgd}
to generate adversarial perturbations for adversarial pre-training and fine-tuning, respectively.

In the pre-training stage,
the augmentations and projection head structure suggested in \cite{chen2020simclr} are adopted.
To optimize the loss function,
we apply SGD (stochastic gradient decent) with weight decay of $5 \times 10 ^{-4}$,
whose learning rate gradually climbs up to 0.1 by linear warmup for the first 10 epochs
and then is decayed by the cosine decay schedule without restarts \cite{loshchilov2017cosine}.

When it comes to fine-tuning on downstream classification tasks,
we take the augmentations popularized by Residual Networks \cite{he2016resnet}:
4 pixels are reflection padded on each side,
and a $32 \times 32$ crop is randomly sampled from the padded image or its horizontal flip.
For full adversarial fine-tuning,
we train the entire network using TRADES \cite{zhang2019trades} for 100 epochs, just like ACL \cite{jiang2020acl}.

\subsection{Pre-training performance}
\label{section-feature-representations}

First, we evaluate the quality of the feature representations learned in pre-training.
To this end, we add a linear classifier on top of the pre-trained query encoder network.
The weights of the classifier are then determined by two ways, one using standard training and the other using adversarial training.
These two ways are denoted as StdEv and AdEv, respectively.
We compare AMOC with the recently proposed ACL \cite{jiang2020acl}.

\begin{figure}[hbt]
	\centering
	\scalebox{0.6}{\includegraphics{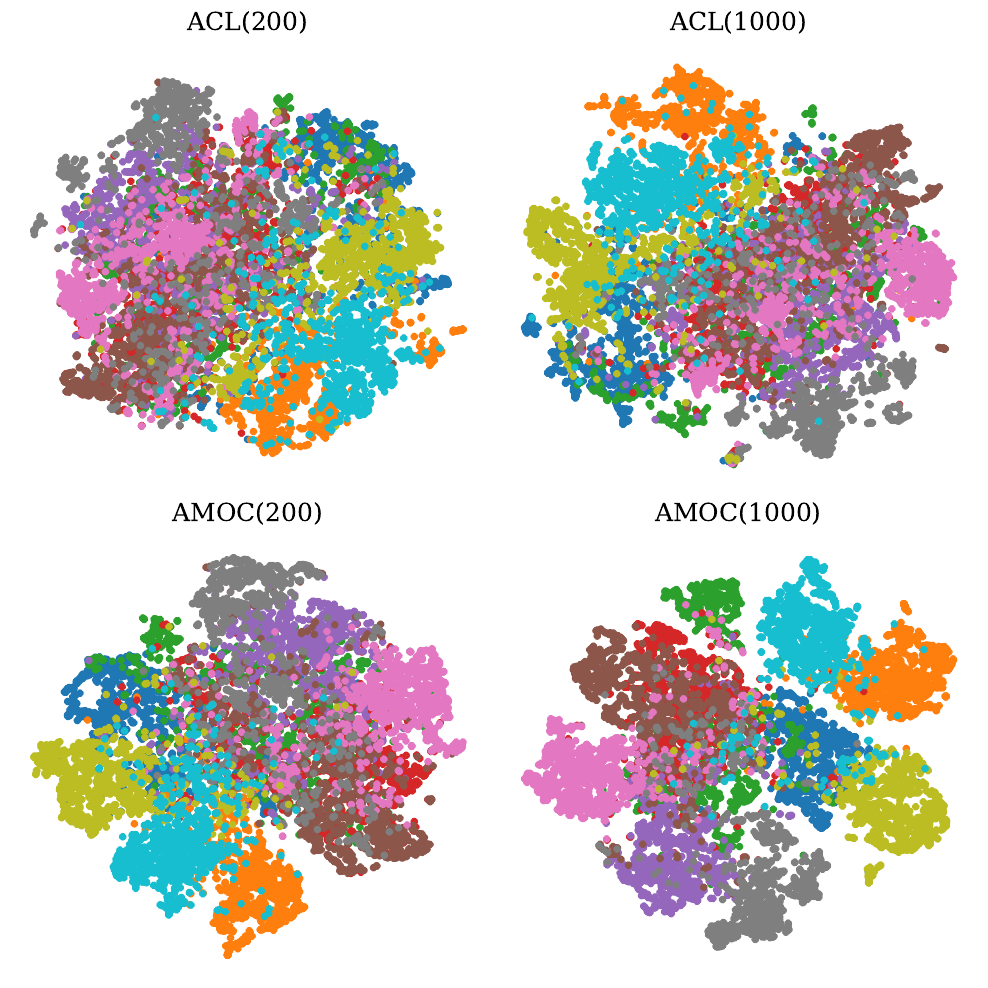}}
	\caption{
		t-SNE \cite{van2008tsne} visualization of the feature representations of ACL \cite{jiang2020acl} and AMOC on CIFAR-10 dataset after 200 and 1000 training epochs.
        Here ACL \cite{jiang2020acl} takes a batch size of 512 while our method only uses a smaller batch size of 256.
		The numbers in ( ) represent the training epochs.
        The plot clearly shows that the feature distribution obtained by our approach has a better separability than that of ACL \cite{jiang2020acl} under a smaller batch size and fewer training epochs.
}
	\label{tsne}
\end{figure}

The clean and robust accuracy are reported in Table \ref{table-ACL-AMOC} using  different batch sizes and training epochs.
As we can see that AMOC consistently outperforms ACL under the same configurations.
In particular, the results of AMOC with a batch size of 64 have far outperformed those of ACL with a batch size of 512.
Moreover, thanks to the constructed memory banks,
our approach converges much faster than ACL.
The performance of AMOC under AdEv using a batch size of 256 after 200 training epochs is comparable to that of ACL using a batch size of 512 after 1000 epochs.
Moreover, it is obvious that MoCo has the best natural accuracy, but hardly any robustness.
Compared to MoCo, our method achieves significant robustness improvements with a small loss of natural accuracy.

\begin{table}[hbt]
	\caption{Comparison of MoCo \cite{he2019moco},  ACL \cite{jiang2020acl} and AMOC  with different batch sizes and epochs on CIFAR10.
	\textbf{StdEv:}  Train the classifier using  \textit{clean} samples through the frozen encoder.
	\textbf{AdEv:}  Train the classifier using  \textit{adversarial} examples through the frozen encoder.
	}
	\label{table-ACL-AMOC}
   	\centering
	\scalebox{0.65}{
	\begin{tabular}{ccccccccc}
		\toprule
		\multicolumn{3}{c}{} && \multicolumn{2}{c}{StdEv} && \multicolumn{2}{c}{AdEv} \\
		\cmidrule{5-6} \cmidrule{8-9}
		&Batch Size&Epochs&&Clean& PGD20&&Clean& PGD20 \\
		\midrule
		\multirow{2}*{MoCo \cite{he2019moco}} & 256 & 200 && 80.44 & 0.17 && 75.27 & 1.36 \\
		& 256 & 1000 && \textbf{90.91} & 0.35 && \textbf{87.00} & 0.50 \\
		\midrule
		\multirow{5}*{ACL \cite{jiang2020acl}} & 64 & 200 && 70.53 & 26.24 && 62.83 & 35.76 \\
		& 128 & 200 && 74.35 & 30.53 && 66.46 & 39.04 \\
		& 256 & 200 && 77.14 & 32.72 && 68.60 & 40.36 \\
		& 512 & 200 && 77.17 & 33.19 && 69.13 & 40.42 \\
		& 512 & 1000 && 82.08 & 40.67 && 74.18 & 44.99 \\
		\midrule
		\multirow{4}*{AMOC} & 64 & 200 && 77.74 & 37.56 && 73.29 & 43.39 \\
		& 128 & 200 && 79.36 & 38.44 && 74.79 & 44.01 \\
		& 256 & 200 && 79.10 & 37.12 && 74.36 & 43.36 \\
		& 256 & 1000 &&86.12 &\textbf{ 45.29} &&81.91 & \textbf{50.28} \\
		\bottomrule
	\end{tabular}}
\end{table}

\begin{figure*}[htb]
	\centering
   	\scalebox{0.7}{\includegraphics{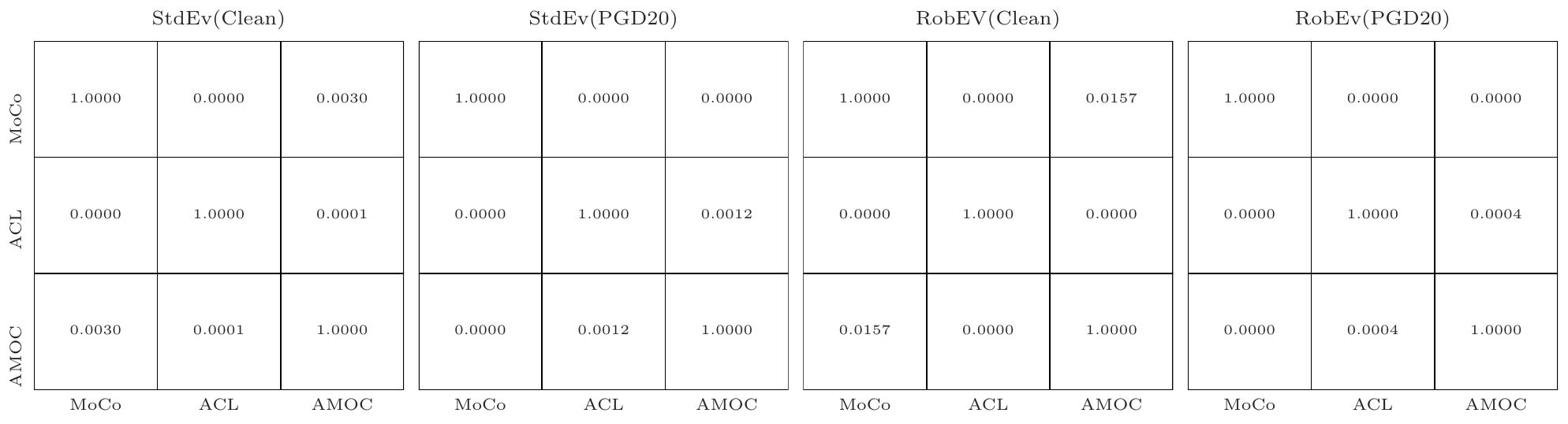}}
	\caption{
		P-values of paired t-tests \cite{rice2006ttest} between  MoCo \cite{he2019moco},  ACL \cite{jiang2020acl} and AMOC  using 5 repetitive pre-training results.
		A lower p-value means a more significant difference in performance between the two methods.
	}
	\label{fig-stats-kcompare}
\end{figure*}

Furthermore,  we conduct the paired t-tests \cite{rice2006ttest} between MoCo, AMOC and ACL based on 5 repetitive pre-training results after 200 epochs.  	
We plot the p-values of statistical analysis under four measures in Figure \ref{fig-stats-kcompare}.
It is obvious that the performance differences are statistically significant under p-value $<$ 0.05.

Next, we compare the running times of ACL and our approach in Table \ref{table-com-cost}.
It is observed that both methods take almost the same running time per epoch.
But as reflected by the previous experimental results,
the performance of AMOC after 200 training epochs is comparable to that of ACL after 1000 epochs,
so the total running time of AMOC could be much less in practice.

\begin{table}[htb]
	\caption{
	The running times of ACL \cite{jiang2020acl}  and AMOC with different batch sizes and epochs on CIFAR10.
	The times are evaluated on an Intel Xeon CPU E5-2680 v4 platform with 16 GB of memory and a single RTX 2080Ti GPU.
	}
	\label{table-com-cost}
	\centering
	\scalebox{0.65}{
	\begin{tabular}{ccccccc}
		\toprule
		&Batch Size&Epochs&& Total Time (hours) & Time per epoch (seconds) \\
		\midrule
		\multirow{3}*{ACL \cite{jiang2020acl}}
       	& 256 & 200 &&  13.90  &  250.15  \\
       	& 512 & 200 &&  13.40  & 241.15   \\
		& 512 & 1000 && 67.01 &  241.15 \\
		\midrule
		\multirow{2}*{AMOC}
       	& 64 & 200 && 15.35  & 276.37 \\
       	& 256 & 200 && 13.33  & 239.94 \\
	   \bottomrule
	\end{tabular}}
\end{table}

\begin{figure*}[hbt]
	\centering
	\scalebox{0.7}{\includegraphics{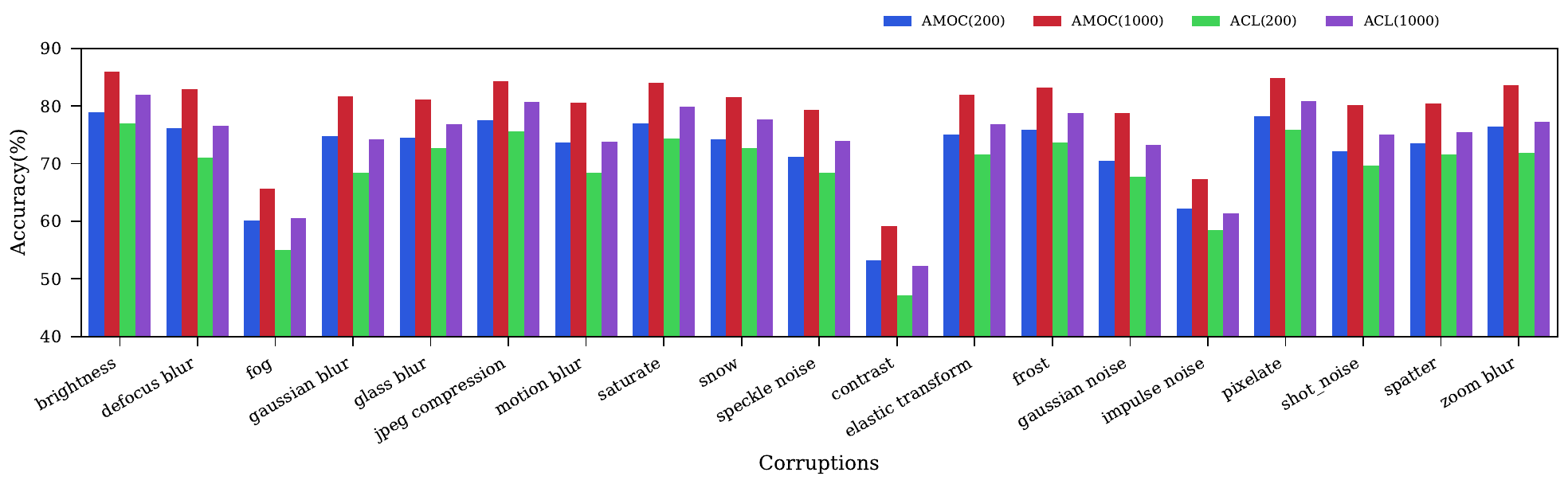}}
	\caption{Comparision of ACL \cite{jiang2020acl} and AMOC against unseen corruptions on CIFAR10-C.
	             The numbers in ( ) represent the training epochs.
	}
	\label{bar-cifar10}
\end{figure*}

Finally, we use the CIFAR10-C dataset \cite{hendrycks2019cifarc} to further evaluate the robustness
of the model against unseen semantic invariant corruptions.
In Figure \ref{bar-cifar10}, we compare the classification accuracy of ACL and AMOC for each type of corruption.
It is obvious that AMOC still  outperforms ACL under different corruptions.
Especially, under some corruptions like gaussian blur and impulse noise,
the advantages of AMOC are particularly pronounced.

\subsection{Comparison with supervised models after fine-tuning}

In this section, we will demonstrate how the encoder network pre-trained by AMOC can further improve the downstream classification tasks.
To this end, we first pre-train the encoder by AMOC,
and then add a linear classifier and  fine-tune the whole network using the adversarial loss of TRADES \cite{zhang2019trades}.
According to the results in the previous section, we choose a batch size of 256 and run 200 epochs in AMOC pre-training.
We compared the fine-tuned results with several supervised defenses,
including PGD-AT \cite{madry2018pgd},
ALP \cite{kannan2018alp} and TRADES \cite{zhang2019trades},  on CIFAR10 and CIFAR100 datasets, respectively.

\begin{table}[htb]
	\center
	\caption{
	Comparison of classification accuracy (\%) between AT, ALP, TRADES and AMOC  under various  adversarial attacks on CIFAR10.
	We take \textbf{bold} type to indicate the best result, and \underline{underline} type to indicate the second best result.
	AMOC: pre-trained results; $\text{AMOC}^{\dagger}$: fine-tuned results.
	}
	\label{table-supervised-cifar10}
	\scalebox{.55}{
		\begin{tabular}{ccccccccccc}
			\toprule
		&	&	 \multicolumn{3}{c}{$\ell_{\infty}$, $\epsilon=8/255$}	& 			&	 \multicolumn{2}{c}{$\ell_1$, $\epsilon=12$}	&	& \multicolumn{2}{c}{$\ell_{2}$, $\epsilon=0.5$}	\\
			\cmidrule{3-5}	\cmidrule{7-8} \cmidrule{10-11}
	&Clean& PGD20 & DeepFool& AutoAttack &  & PGD50 & SLIDE & & PGD50 & C\&W\\
			\midrule
Standard Training &\textbf{95.28}&0.00&0.03&0.00&&6.60&3.21&&0.34&0.03\\
PGD-AT \cite{madry2018pgd} &84.74&45.50&50.30&41.57&&54.93&21.33&&54.76&54.48\\
ALP \cite{kannan2018alp} &\underline{85.07}&49.09&\textbf{54.59}&44.13&&57.55&23.03&&57.19&56.32\\
TRADES \cite{zhang2019trades} &81.11&\underline{51.89}&51.89&\underline{47.07}&&58.20&\underline{27.84}&&\underline{59.53}&\underline{57.17}\\
\midrule
AMOC &74.36&43.46&40.85&34.61&&\underline{58.47}&\textbf{31.92}&&55.10&50.98\\
$\text{AMOC}^{\dagger}$&82.76&\textbf{53.51}&\underline{54.45}&\textbf{48.75}&&\textbf{59.85}&27.82&&\textbf{60.57}&\textbf{58.41}\\
\bottomrule
\end{tabular}}
\end{table}

\begin{table}[htb]
	\center
	\caption{
	Comparison of classification accuracy (\%) between PGD-AT, ALP, TRADES and AMOC  under various  adversarial attacks on CIFAR100.
	We take \textbf{bold} type to indicate the best result, and \underline{underline} type to indicate the second best result.
	AMOC: pre-trained results; $\text{AMOC}^{\dagger}$: fine-tuned results.
	}
	\label{table-supervised-cifar100}
	\scalebox{.55}{
		\begin{tabular}{ccccccccccc}
			\toprule
		&	&	 \multicolumn{3}{c}{$\ell_{\infty}$, $\epsilon=8/255$}	& 			&	 \multicolumn{2}{c}{$\ell_1$, $\epsilon=12$}	&	& \multicolumn{2}{c}{$\ell_{2}$, $\epsilon=0.5$}	\\
			\cmidrule{3-5}	\cmidrule{7-8} \cmidrule{10-11}
	&Clean& PGD20 & DeepFool& AutoAttack &  & PGD50 & SLIDE & & PGD50 & C\&W\\
			\midrule
Standard Training &\textbf{76.34}&0.03&0.02&0.00&&1.92&1.20&&0.23&0.64\\
PGD-AT \cite{madry2018pgd} &56.48&20.98&22.65&18.83&&29.62&9.06&&28.70&28.28\\
ALP \cite{kannan2018alp} &55.71&25.97&\underline{25.32}&\underline{22.02}&&32.93&13.18&&32.38&30.21\\
TRADES \cite{zhang2019trades} &54.95&\underline{26.72}&24.39&21.70&&\underline{35.47}&\textbf{15.84}&&\underline{34.26}&\underline{31.24}\\
\midrule
AMOC &42.11&20.50&14.95&11.90&&31.27&14.78&&28.57&22.87\\
$\text{AMOC}^{\dagger}$ &\underline{57.64}&\textbf{28.90}&\textbf{26.88}&\textbf{24.08}&&\textbf{36.66}&\underline{15.64}&&\textbf{35.71}&\textbf{32.78}\\
\bottomrule
\end{tabular}}
\end{table}

It can be found in Table \ref{table-supervised-cifar10} that
AMOC with fine-tuning ($\text{AMOC}^{\dagger}$) surpasses other defenses under most attacks.
Although PGD-AT and ALP can achieve better natural accuracy,
they perform poorly under AutoAttack, 7\% and 4\% less than ours.
Recall that AMOC employs the same adversarial loss as TRADES for fine-tuning,
but AMOC outperforms TRADES by 1.5\% regardless of natural accuracy or robustness.
It indicates that the encoder pre-trained using AMOC is beneficial to
improve the robustness while preserving high natural accuracy.

Similar trends can also be observed on CIFAR100 as shown in Table \ref{table-supervised-cifar100}.
AMOC is not only robust against most attacks,
but also outperforms other defense methods on natural accuracy.
Therefore, adversarial training on top of the pre-trained encoders
is arguably a good choice for efficiency and robustness purposes.

\begin{figure}[htb]
	\centering
	\begin{minipage}{0.22\textwidth}
			\centering
			\scalebox{.7}{\includegraphics{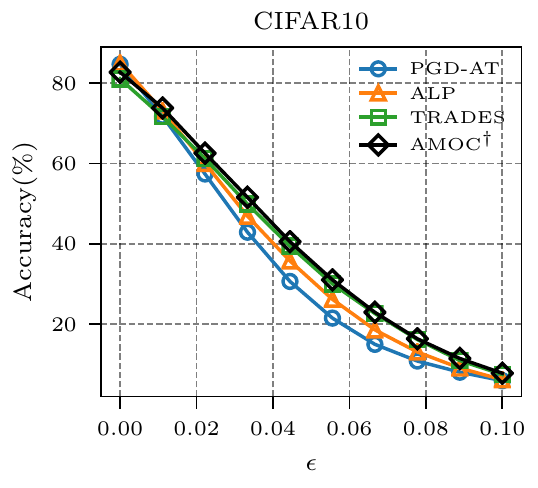}}
		\end{minipage}
	\begin{minipage}{0.22\textwidth}
			\centering
			\scalebox{.7}{\includegraphics{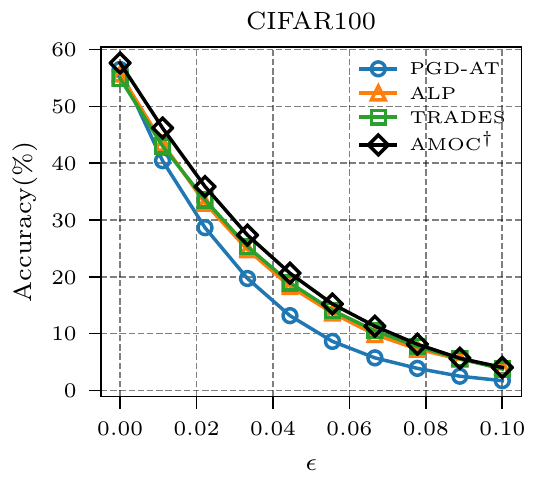}}
		\end{minipage}
	\caption{Comparison between PGD-AT, ALP, TRADES and AMOC under PGD20 attacks with  various perturbation budgets $\epsilon$.}
	\label{epsilon}
\end{figure}

Although all defense methods are adversarially trained within
$\epsilon = 8/255$, it would be better if they could extrapolate to lager perturbations.
Figure \ref{epsilon} depicts the relationship between classification accuracy and increasing perturbation budgets
under PGD20 attack.
Compared with PGD-AT and ALP, AMOC maintains stable and extraordinary robustness,
which demonstrates the reliability and scalability of our approach.

\subsection{Other variants}
\label{AMOCv}
Note that in the second loss term $\mathcal{L}_{\mathrm{ACA}}$ of \eqref{FinalLoss},
we inject \textbf{A}dversarial perturbations into the query encoder,
using the key encoding of \textbf{C}lean augmented data to provide the positive key
and the \textbf{A}dversarial memory bank to provide the negative keys.
If we replace the loss term $\mathcal{L}_{\mathrm{ACA}}$ with $
\mathcal{L}_{\mathrm{ACC}}:=\mathcal{L}_{\mathrm{NCE}}(f_q(c(x)+\delta;\mathrm{BN_{adv}}), f_k(c'(x);\mathrm{BN_{clean}}), \mathcal{M}_{\mathrm{clean}}),
$
then we can derive another pre-training version whose optimization objective is $\lambda \mathcal{L}_{\mathrm{CCC}} + (1- \lambda) \mathcal{L}_{\mathrm{ACC}}$.
For simplicity, we call this version ACC.
It differs from original AMOC (ACA)  in that only a single clean memory bank is used to provide negative keys.
However, as can be seen from the  first part of  Table \ref{table-combinatons},
although this version shows good robustness and natural accuracy,
its performance is worse than that of ACA,
which demonstrates that the newly proposed adversarial memory  is indeed essential to improve the performance of the model further.

Some people might consider feeding adversarial samples to the key encoder.
However, this is not a good choice because it will yield vague positive keys, making the model difficult to learn.
We provide the corresponding verification in the second part of Table \ref{table-combinatons}.
It is observed that injecting perturbations into the key encoder  can greatly deteriorate the performance of the model.

\begin{table}[!hbt]
	\renewcommand{\arraystretch}{1.}
	\caption{Configurations of different variants of AMOC.
                                 \Checkmark means injecting the adversarial perturbations to the corresponding encoder while \XSolidBrush  means not injecting the perturbations.}
	\label{combination}
	\centering
	\scalebox{.8}{
		\begin{tabular}{c|cc|cccc}
	\hline
	& ACA	& ACC	&  AAA	& AAC  &  CAA	& CAC\\
	\hline
	
Query encoder         & \Checkmark	& \Checkmark	&  \Checkmark	 & \Checkmark  &  \XSolidBrush	& \XSolidBrush \\
Key encoder            &  \XSolidBrush	& \XSolidBrush	& \Checkmark & \Checkmark & \Checkmark & \Checkmark\\
Memory bank 	& $\mathcal{M}_{\mathrm{adv}}$	& $\mathcal{M}_{\mathrm{clean}}$ 	& $\mathcal{M}_{\mathrm{adv}}$	& $\mathcal{M}_{\mathrm{clean}}$ & $\mathcal{M}_{\mathrm{adv}}$	& $\mathcal{M}_{\mathrm{clean}}$  \\
	\hline
	\end{tabular}}
\end{table}

\begin{table}[htb]
	\caption{Comparison of classification accuracy (\%) of possible variants on CIFAR10.
                \textbf{StdEv:}   Train the classifier using  clean samples with frozen encoder.
                \textbf{AdEv:}  Train the classifier using  adversarial examples with frozen encoder.}
    \label{table-combinatons}
   	\centering
	\scalebox{0.9}{
		\begin{tabular}{ccccccc}
			\toprule
		& &\multicolumn{2}{c}{StdEv}	& &\multicolumn{2}{c}{AdEv}		\\
			\cmidrule{3-4}	 \cmidrule{6-7}
	    & &Clean & PGD20 & & Clean & PGD20   \\
\hline
ACA & &\underline{79.10}  &\textbf{37.12} && \textbf{74.36} & \textbf{43.36} \\
ACC  & &78.74  &\underline{36.87} && \underline{74.08} & \underline{42.89} \\
\hline
AAC  & &10.00  &10.00 && 10.00  & 10.00 \\
AAA  & &68.52 &34.44  &&  63.66 & 39.20 \\
CAC  & &78.46  &0.00  && 24.82  & 15.53  \\
CAA  & &\textbf{79.29}  &0.00  && 27.12  & 15.08 \\
\bottomrule
	\end{tabular}}
\end{table}

\subsection{Sensitivity analysis}
\label{section-options}
\begin{figure}[htb]
	\centering
	\scalebox{.85}{\includegraphics{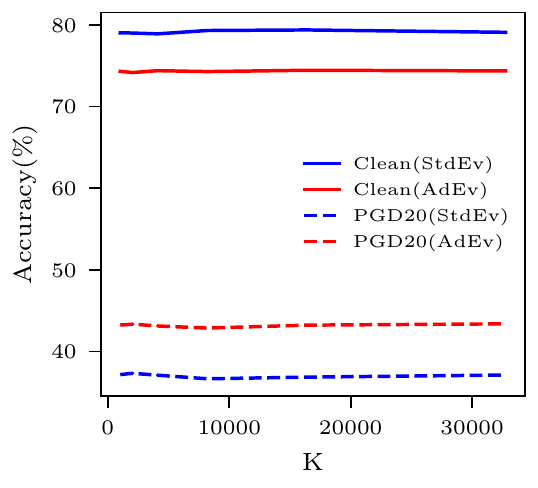}}
	\caption{The performance of AMOC across different length of memory banks.
                \textbf{StdEv:}   Train the classifier using  clean samples with frozen encoder.
                \textbf{AdEv:}  Train the classifier using  adversarial examples with frozen encoder.
	}
	\centering
	\label{ablation_K}
\end{figure}

The length of the memory banks decides the amount of the historical information used in each training epoch.
We investigate its influence by considering  the robustness and natural accuracy under various length choices.
As shown in Figure \ref{ablation_K},
the performance of the our approach is not sensitive to the length of the memory banks,
and both robustness and natural accuracy are very stable as $K$ changes.

\begin{figure}[htb]
	\centering
	\scalebox{.75}{\includegraphics{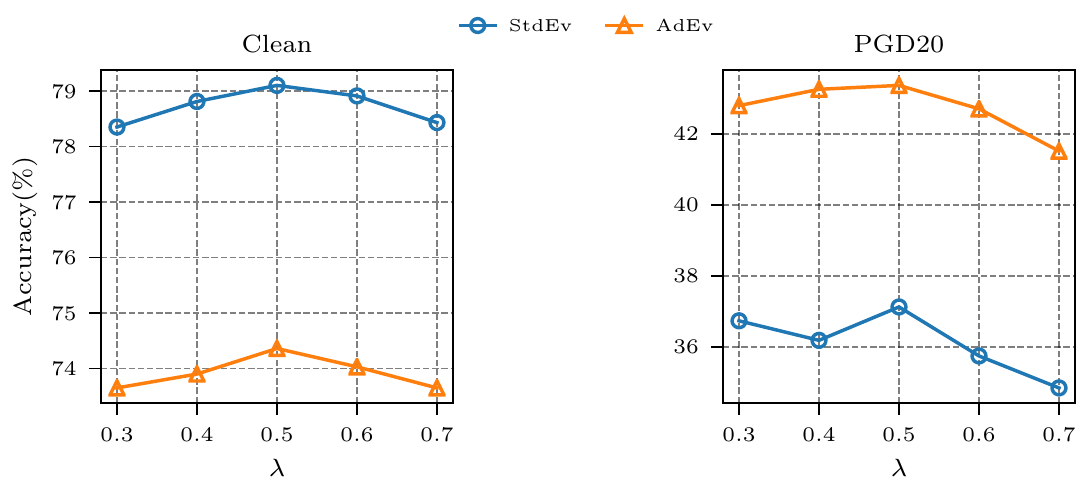}}
	\caption{The clean accuracy and PGD20 robustness of AMOC with respect to different balance hyperparameter $\lambda$.
                \textbf{StdEv:}   Train the classifier using  clean samples with frozen encoder.
                \textbf{AdEv:}  Train the classifier using  adversarial examples with frozen encoder.
	}
	\centering
	\label{ablation_lambda}
\end{figure}

The balance hyperparameter $ \lambda $ in  \eqref{FinalLoss} controls the effects of the adversarial samples and clean samples.
We plot the natural accuracy and robustness with respect to different values of $ \lambda $ in Figure \ref{ablation_lambda}.
It is observed that either too large or too small $\lambda$ can deteriorate the performance of the model,
which means that the clean loss $\mathcal{L}_{\mathrm{CCC}}$  and adversarial loss $\mathcal{L}_{\mathrm{ACA}}$ are equally important in our framework.
Moreover, a too large $\lambda$ seems more damaging to robustness,
which implies that overemphasizing clean samples can have a larger negative impact.

\section{Conclusion}
In this work, we incorporate adversarial perturbation into the perspective of data augmentation,
and propose a simple but efficient momentum-based contrastive pre-training to improve the robustness and natural accuracy of the neural networks.
Thanks to the designed memory banks,
compared to the previous adversarial self-supervised learning approach,
the proposed model can learn more discriminative feature representations using a smaller batch size and far fewer epochs.
However, since we still rely on adversarial attacks to yield augmented samples in the training,
the approach still bears a heavy computational burden.
How to further accelerate the 
pre-training is a promising direction in the future.

\end{document}